# Conversational Structure Aware and Context Sensitive Topic Model for Online Discussions

Yingcheng Sun
EECS Department
Case Western Reserve University
Cleveland, Ohio, USA
Email: yxs489@case.edu

Kenneth Loparo
EECS Department
Case Western Reserve University
Cleveland, Ohio, USA
Email: kal4@case.edu

Richard Kolacinski
EECS Department
Case Western Reserve University
Cleveland, Ohio, USA
Email: rmk4@case.edu

*Abstract*—Millions of online discussions are generated everyday on social media platforms. Topic modelling is an efficient way of better understanding large text datasets at scale. Conventional topic models have had limited success in online discussions, and to overcome their limitations, we use the discussion thread tree structure and propose a "popularity" metric to quantify the number of replies to a comment to extend the frequency of word occurrences, and the "transitivity" concept to characterize topic dependency among nodes in a nested discussion thread. We build a Conversational Structure Aware Topic Model (CSATM) based on popularity and transitivity to infer topics and their assignments to comments. Experiments on real forum datasets are used to demonstrate improved performance for topic extraction with six different measurements of coherence and impressive accuracy for topic assignments.

*Keywords- Online discussions; Topic modeling; Conversational structure*

## I. INTRODUCTION

With the prevalence of content sharing platforms, such as online forums, microblogs, social networks, photo and video sharing websites, people are more and more accustomed to expressing and sharing their opinions on the Internet. Modern news websites provide commenting facilities for their readers to freely post and reply. The increasing popularity of such platforms results in huge amounts of online discussions each day. Automatically modeling topics from massive texts can help people better understand the main clues and semantic structures, and can also be useful to downstream applications such as discussion summarization [9], stance detection [6], event tracking [8], and so on.

Conventional topic models, like probabilistic Latent Semantic Analysis (pLSA) [10] and Latent Dirichlet Allocation (LDA) [2] assume documents have latent semantic structure ("topics") that can be inferred from word–document co-occurrences. They have achieved great success in modeling long text documents over the past decades, but may not work well when directly applied to short texts that dominate online discussions for two reasons about the data: 1) **Sparse**: The occurrences of words in short documents have a diminished discriminative role compared to lengthy documents where the model has sufficient word counts to determine how words are related. [11] 2) **Noisy**: Comment threads often contain unproductive banter, insults, and cursing, with users often "shouting" over each other [20], and people sometimes publish "unserious" response posts that are unrelated to the discussion topics [5]. Noisy comments perhaps could be used for sentiment analysis, but are significant disturbances when extracting topics from discussion threads.

To address the issues discussed above, in this paper, we use the tree structure that each discussion thread inherently exhibits based on the relationship between postings and replies to enrich the background information of each comment. Fig. 1 illustrates a typical discussion thread of user comments on a submitted question and its corresponding tree structure.

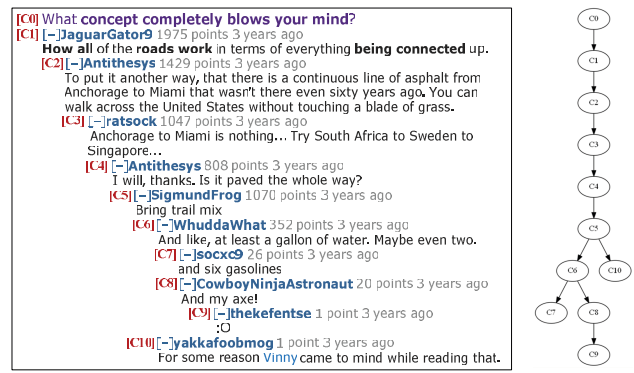

Figure 1. An example thread of user comments on the posted question: "the concept completely blows your mind"[1] with the original nested discussion on the left and its corresponding Tree structure on the right.i: the i-th comment

In Fig. 1, the occurrence frequency of each word in the possible topic "concept of 'how all roads work' completely blows your mind" equals to or even less than those "non-topical" words, making it very difficult to be modeled using conventional topic models. However, we can see that different comment nodes have different numbers of replies, and nodes (node 0 and 1) leading the topics have more replies than others, and those nodes are also in relatively "higher" positions in the discussion tree, above their topic "following" nodes. Motivated by this observation, we propose "popularity" metric to measure the number of replies to a comment as an extension to the frequency of word occurrence. We also observe that the topic distribution of a node is dependent on its parent because comments in reply to the content of their parents form a

---

[1]https://www.reddit.com/r/AskReddit/comments/3dtyke/what\_concept\_completely\blows\_your\_mind

85

conversational thread. We use this "transitivity" characteristic as context information to reduce the inaccuracy of topic assignments to comments, especially for those "noisy" ones, like comment 9 in Fig. 1. Based on the above two characteristics, we build a Conversational Structure Aware Topic Model (CSATM) that makes the topics modeled meaningful and usable, and robust to noisy comments.

## II. RELATED RESEARCH

Topic models aim to discover latent semantic information, i.e., topics, from texts and have been extensively studied. Latent Dirichlet Allocation [2] is a widely used topic model that represents a document as a mixture of latent topics to be inferred, where a topic is modeled as a multinomial distribution of words. Nevertheless, prior research has demonstrated that topic models only focusing on word–document co-occurrences are not suitable for short and informal texts like Tweets, reviews, and online comments due to data sparsity and noise [30, 48, 49]. Therefore, three main strategies are proposed by recent researchers to tackle these problems and we provide a brief overview of them.

### A. Merging Shorts Texts into Long Pseudo Documents

The idea of this strategy is merging related short texts together and applying standard topic modeling techniques on the pooled documents. Auxiliary contextual information is used during the merging process, like authors, time, locations, hashtags, conversations, and etc. For example, Weng et al. [33], Hong and Davison [11], and Zhao et al. [36] heuristically aggregate messages posted by the same user or that share the same words before conventional topic models are applied. Alvarez-Melis and Saveski [1] group tweets together occurring in the same user-to-user conversation. Ramage, Dumais, and Liebling [27] and Mehrotra et al. [18] employ hashtags as labels to train supervised topic models. The performance of these models can be compromised when facing unseen topics that are irrelevant to any hashtag in the training data.

In practice, auxiliary information is not always available or just too costly for deployment, so models without using auxiliary information have been put forward, like Self-Aggregation-based Topic Model (SATM) [26], Pseudo-document-based Topic Model (PTM) [37], and etc. However, those models still could not deal with the case when the data is extremely sparse and noisy like the example Fig. 1 shows, and no prior knowledge is given to ensure the quality of text aggregation, that will further affect the performance of topic inference.

### B. Building Internal Relationships of Words

This strategy uses the internal semantic relationships of words to overcome the problem of lacking word co-occurrence, and the semantic information of words has been effectively captured by deep-neural network-based word embedding techniques. Several attempts [29, 34] have been made to discover topics for short texts by leveraging semantic information of words from existing sources. These topic models rely on a meaningful embedding of words obtained through training on a large-scale high-quality external corpus, which should be both in the same domain and language as the data used for topic modeling.

However, such external resources are not always available [42-47]. The SeaNMF [30] model learns the semantic relationship between words and their context from a skip-gram view of the corpus. The Biterm Topic Model (BTM) [35] and the RNN-IDF-based Biterm Short-text Topic Model (RIBSTM) [17] model biterm co-occurrences in the entire corpus to enhance topic discovery. Latent Feature LDA (LFTM) [23] incorporates latent feature vector representations of words. The relational BTM model (R-BTM) [15], links short texts using a similarity list of words computed using an embedding of the words. However, because social media content and network structures influence each other, only focusing on content is insufficient [50, 51].

### C. Leveraging Discussion Tree Structure as Prior

The third line of research focuses on enriching prior knowledge when training the topic model. LeadLDA [13] distinguishes reply nodes into "leaders" and "followers" in the conversation tree, and models the distribution of topical and non-topical words from "leaders" and "followers", respectively. To detect "leaders" and "followers" in the tree structure, the first step is to extract all root-to-leaf paths and then classifying nodes in each path using a supervised learning model after labeling, and then combing all paths [12]. Extracting and combing paths is time consuming and labeling is labor intensive, so LeadLDA may not be suitable for large online discussion datasets. Li et al. [14] exploits discourse in conversations and joins conversational discourse and latent topics together for topic modeling. This model also organizes microblog posts as a conversation tree structure, but does not consider topic hierarchies and model robustness issue like our proposed model.

Hierarchical Dirichlet Process (HDP) [31] and Nested Hierarchical Dirichlet Process (nHDP) [24] can build hierarchical topic models with nonparametric Bayesian networks, but they model the hierarchical structure of topics, not the documents. In online discussions, if we treat each comment as a document, the comment it replies to and its following replies all provide plentiful clues for its topic inference, which is not discussed in HDP or nHDP. Learning based methods are also explored [52 -54]. In this paper, we will introduce a model that uses the conversational structure [39, 40, 41] of a discussion thread inherently has to improve the topic modeling performance for short texts within online discussions.

## III. CONVERSATIONAL STRUCTURE AWARE TOPIC MODEL

Our model extends the LDA model by adding the structural relationships among nodes in a discussion tree as context information for each online comment. With the conversational structure, we observe the "popularity" and "transitivity" characteristics of topics in online discussions. We will introduce the intuitions on "popularity" and "transitivity" and how we use them in our model to make extracted topics meaningful and usable.

### A. Topic Generation with Popularity

In online discussions, users can easily participate by submitting comments or writing replies to those that draw their attention. In writing a reply, a user reads the initial post or headline, browses the comments and selects one for a reply. By

86

writing a reply, a user explicitly expresses their interest in the topic(s) in the discussion thread, thereby increasing their popularity and enlarging the discussion tree by adding leaf nodes. The main topics of a reply may not be closely related to comments located at a distance in the discussion thread, but will definitely be responsive to the comment it is directly replying to. We thus design our model based on two intuitions:

1) **The popularity of topics discussed in a comment node is positively related to the number of replies.**
2) **The topic distribution of a node is dependent on its ancestors, and the dependency is negatively related to the distance from the node to its ancestor.**

For intuition 1), the word "popularity" is commonly used as the state or condition of a person or item being liked by the people. The popularity of an item usually depends on the number of people that support it. As the readers to a book and the audience to a movie, the popularity of a topic can be measured by the number of people that are involved in its discussions. There may be various reasons that a topic becomes popular like its creation time, the celebrity of its author or the topic itself, but the reasons are not what we are going to discuss in this paper. We are more interested in finding the most popular and influential topics in an online discussion thread, and we also believe that such kind of topics should be extracted by topic models. As the discussion tree example Fig. (2 a) shows, root node 1 may put forward a main topic with three replies: nodes 2, 3, and 4. If we assume these three nodes discuss three "sub-topics", then the sub-topic in node 3 is the most popular because it receives the most responses and should be assigned with higher possibility.

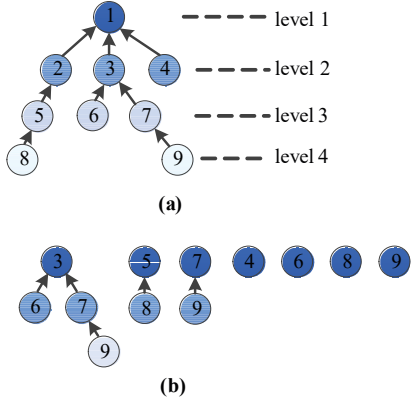

Figure 2. (a) Example of a discussion tree with 4 levels (b) Subtrees used for calculating popularity scores of nodes 2 to 9. Shade of color represents topic "influence" of the root, the deeper the stronger the influence.

Following intuition 1), the "popularity" $p_i$ of node $i$ depends on all replies in its subtrees, and replies in different level have different weights but the same weight in the same level; so $p_i$ can be written as:

$$p_i = \sum_l \sum_{n_l} w_l * node = \sum_{d_i} w_l * p_j \qquad (1)$$

where $n_l$ is the number of nodes in level $l$, and $w_l$ is the weight for nodes in level $l$. We can also write the popularity score of a node as the sum of its children's popularity scores by iterative accumulation, and $d_i$ is the degree of node $i$. We need to be careful that all counts should be taken in node $i$'s subtree. As Fig. (2 b) shows, node 2's popularity is calculated only on node 5 and 8, not on any other node. Also, we set the initial popularity of any node as 1 in this paper, so the popularity value of a node without any reply is 1 that is its initial value, like node 4, 6, 8 and 9 in Fig. 2. For nodes with replies, like node 1, 2, 3, 5 and 7, their popularity values are the sum of the initial popularity and the popularity of replies. According to intuition 2), the popularity of replies in different levels does not have the same weight.

For intuition 2), let's assume there is a comment node $i$ in the discussion tree $t$. Users can choose any comment to reply in $t$, but if $i$ is chosen, it indicates that the topics in comment node $i$ attract the users more than other nodes. The newly added child node to $i$ continue the topics discussed in $i$, making topic transitive from $i$ to its children, but the topic shift [32, 38] and the topic drift [16, 25] phenomenon make the transitivity process with some "loss", so the "topic influence" of a root decreases when the discussion thread gets longer. In Fig. (2 a), the topic introduced in node 1 spreads across the entire tree, but its influence will weaken from level 1 to level 4 because of the topic transitivity loss. We thus use a decreasing sequence to model the weight $w_l$ in equation (1) and we assume that nodes in level $l$ of the subtree have the same weight. We list three different options as the decreasing sequence:

a) Arithmetic progression
$$w_l = c - (l-1)d$$
b) Geometric progression
$$w_l = cr^{l-1}$$
c) Harmonic progression with "Gravity" power
$$w_l = (c + (l-1)b)^{-G}$$

where $c$ is a constant, $d$ is the common difference for arithmetic progression, $l$ is the number of the level, and $r$ is the common ratio for the geometric sequence. $G$ is the "gravity" power controlling the fall rate of weights for harmonic progression, and the weight decreases faster the larger $G$ is. If $G=1$, it becomes general harmonic series, where $c$ and $b$ are real numbers. From arithmetic progression to harmonic progression, the weigh distribution curve will become smoother. Fig. 3 shows their differences.

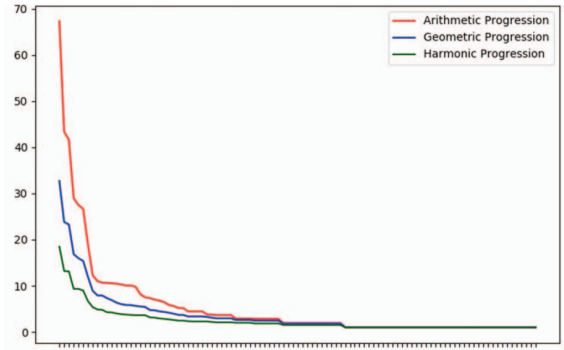

Figure 3. Distributions of popularity scores calculated by arithmetic, geometric and harmonic progressions on the same datasets.

The distribution of popularity score computed by arithmetic progression is sharper, meaning that nodes leading a discussion with a large number of descendants will be given more weights than the other two, so if the dataset is very sparse or topical words are corrupted by noises, the arithmetic progression will be a better choice. From arithmetic progression to harmonic progression, the weight distribution curve becomes smoother and smoother. The choice of sequence is based on the word distribution of datasets, and other sequence can also be used if it fits the modeling requirements.

*B. Model Inference*

CSATM extends the LDA model by integrating the popularity property for each online comment. The latent variables of interest are the topic assignments for work tokens $z$, the comment level topic distribution $\theta$ and the topic – word distribution $\Phi$. The multinomial distribution $\theta$ and $\Phi$ can be efficiently marginalized due to the conjugate Dirichlet-multinomial design, we thus only need to sample the topic assignments $z$. It is computationally intractable to compute the exact posterior distribution using Gibbs sampling for approximating inference. To perform Gibbs sampling, we first choose initial states for the Markov chain randomly. Then we calculate the conditional distribution $p(z_i = k | z^{-i}, w, p_c, \alpha, \beta)$ for each word, where the superscript '$-i$' signifies leaving the $i$th token out of the calculation, $w$ is the global word set, and $p_c$ is the popularity score for comment $c$. By applying the chain rule on the joint probability of the data, we can obtain the conditional probability as:

$$p(z_i = k | z^{-i}, w, p_c, \alpha, \beta) \propto (n_{k,c}^{-i} \lambda\, p_c + \alpha_k) \frac{n_{k,w}^{-i} \lambda\, p_c + \beta_w}{\sum_w n_{k,w}^{-i} \lambda\, p_c + \beta_w}$$

where $n_{k,c}$ is the number of words in comment $c$ that are assigned to topic $k$, and $n_{k,w}$ is the number of times that topic $k$ is assigned to word term $w$, both of which are scaled by the popularity score, and $\lambda$ is the scaling ratio. Following the conventions of LDA, here we use symmetric Dirichlet priors $\alpha$ and $\beta$. Based on the topic assignments of word occurrences, we can estimate the topic-word distributions $\phi$ and global topic distributions $\theta$ as:

$$\phi_{k,w} = \frac{\beta_w + n_{k,w} \lambda\, p_c}{\beta_w + \sum_w n_{k,w} \lambda\, p_c}, \quad \theta_{k,c} = \frac{\alpha_w + n_{k,c} \lambda\, p_c}{\alpha_w + \sum_k n_{k,c} \lambda\, p_c}$$

*C. Topic Assignment with Transitivity*

After discovering usable topics from the corpus, we want the correspondence of topic assignments to documents to be meaningful. Conventional topic assignment methods do not consider the document context information, because for most of the corpus, documents are not dependent. However, comments in online discussions demonstrate clear topic dependency through their nested reply relationships, so we propose a new topic assignment strategy. With CSATM, we obtain the topic distribution for each given comment, and then work out new topic assignments for the comments using the topic transitivity property:

$$t'_i = \frac{\sum_{j=1}^{l_i} w_{l_i - j + 1} t_i^j}{\sum_{j=1}^{l_i} w_{l_i - j + 1}}, \quad i = 1 \dots N$$

where $t'_i$ is the new topic assignment compared to the original assignment $t_i^j$ for comment $i$, and $j$ is the relative order in the path from comment node $i$ to the root, and $l_i$ is the level where node $i$ is located, and $w$ is the weight of level $l_i$ used for calculating the popularity score.

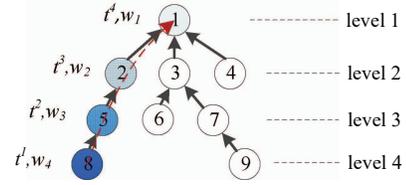

Figure 4. Topic assignment using the topic "transitivity" property in a discussion tree, determining the topic distribution of node 8. The shades of color represent topic dependency, the deeper the color the greater the dependency, with white representing no dependency.

In Fig. 4, the topic distribution of node 8 depends on that of nodes in its path to the root, which are nodes 5, 2 and 1, and does not depend on any node out of the path to the root in terms of the topic distribution. The dependency weakens as the level increases because comments indicate stronger interests in their parent nodes they reply to in upper level than nodes in other levels as discussed intuition 2). By using this new strategy, we can reduce the inaccuracy and uncertainty when assigning topics to noisy comments.

## IV. EXPERIMENT

In this section, we evaluate the proposed CSATM against LDA and several state-of-the-art baseline methods on two real world datasets. We report the performance in terms of six different coherence measures, and compare the accuracy for topic assignments.

*A. Datasets, Compared Models, and Parameter Settings*

In the experiment, we use the Reddit dataset. Reddit [2] is an online discussion website. Registered members can submit content to the site such as links, text posts, or images, and write comments or reply other comments. Posts are organized by subject into user-created boards called "subreddits", which cover a variety of topics. The dataset is obtained from a data collection forum containing 1.7 billion messages (221 million conversations) from December 2005 to March 2018 [3].

After prepossessing, we find that there are 42% posts without any comments and 35% posts with less than or equal to 5 comments. Most of these discussions only focus on one rather than multiple topics and do not have the topic shift phenomenon, so their topics are easy to be modeled accurately, or we can just use the title of each discussion thread as its topic. In order to prove the effectiveness of our proposed model, we thus filter the posts with the number of replies less than 100, and then randomly picked 200 discussions from 30 different "subreddits". No category information is available for this dataset, so three annotators were asked to label each conversation with the topics, and labels agreed by at least two annotators are used as the ground truth, with a total of 810 topics labeled in this manner. We use a web-based text annotation tool called "Tagtog" to annotate the topics for each discussion, as Fig. 5 shows. During the annotation process, the number of topics needs to be set first, and topic assignment of each comment needs to labeled, but the topic set is automatically generated and updated as the labeling work goes on. In addition, the annotation tool will find all the same words across the document and label them, so annotators only need to focus on the words that have not been labeled. In Fig. 5, the labeled words are marked different colors by topics. To simplify the labeling and topic modeling process, each comment is assigned only 1 topic, and the discussion thread is labeled 4 topics on average to avoid too detailed topic assignment.

---
[2] https://www.reddit.com/
[3] https://files.pushshift.io/reddit/

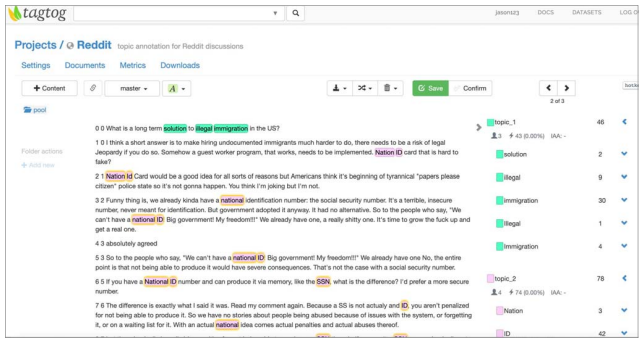

Figure 5. An example of topic annotation interface of Tagtog.

We evaluate the performance of the following models, using all their original implementations.

- LDA: The classic Latent Dirichlet Allocation (LDA) model is used as the baseline model. For every dataset, the LDA model is used by setting the hyper parameters $\alpha = 0.1$ and $\beta = 0.01$, and the number of topics = 70.

- PTM: Pseudo document based Topic Model [37] aggregates short texts against data sparsity. The original implementation with the number of pseudo documents = 1000 and $\lambda = 0.1$.

- BTM: Biterm Topic Model [35] directly models topics of all word pairs (biterms) in each post and explicitly models the word co-occurrence patterns to enhance topic learning. Following the original paper, $\alpha = 50/K$ and $\beta = 0.01$.

- LeadLDA: Generates words according to topic dependencies derived from conversation trees [13]. A classifier trained to differentiate leader and follower messages is required before using LeadLDA [12], labelled leader and follower messages and CRF are used to obtain the probability distribution of leaders and followers.

- LFTM: Latent Feature LDA [23] incorporates latent feature vector representations of words trained on very large corpora to improve the word-topic mapping learnt on a smaller corpus. Following the paper, the hyper-parameter $\alpha = 0.1$.

- SATM: Self-Aggregation-Based Topic Model [26] aggregates documents and infers topics simultaneously. Following [13], the pseudo-document number is chosen from 100 to 1000 in all evaluations, and the best scores are reported.

- CSATM: We need to select a decreasing sequence to model the weights of the levels used for calculating the popularity score. In this experiment, we use the arithmetic progression with the "sharper" weight distribution because the word distribution of the dataset is pretty sparse and 74% of words show up only once.

B. Coherence Evaluation

Topic model evaluation is inherently difficult. In previous work, perplexity is a popular metric to evaluate the predictive abilities of topic models using a held-out dataset with unseen words [2]. However, Chang et al. [4] have demonstrated that the method does not translate to the actual human interpretability of topics, so the coherence score is widely used to measure the quality of topics [26], assuming that words representing a coherent topic are likely to co-occur within the same document [37]. To reduce the impact of low frequency counts in word co-occurrences, we employ the topic coherence metric called normalized pointwise mutual information (PMI, NPMI) [3]. Given the $T$ most probable words in a topic $k$, *NPMI* is computed by:

$$NPMI(k) = \frac{2}{T(T-1)} \sum_{1 \leq i \leq j \leq T} \frac{log \frac{p(w_i, w_j)}{p(w_i)p(w_j)}}{-\log p(w_i, w_j)}$$

where $p(w_i)$ and $p(w_i, w_j)$ are the probabilities that word $w_i$ occurs, and that the word pair $(w_i, w_j)$ co-occurred estimated by the reference corpus, respectively. $T$ is set to 10 in our experiments. We also use five other confirmation measures to further enhance the comparisons across models.

$C_{UCI}$ is a coherence that is based on a sliding window and the PMI of all word pairs of the given top words [21]. The word co-occurrence counts are derived using a sliding window with the size 10. For every word pair, the PMI is calculated. The arithmetic mean of the PMI values is the result of this coherence. $C_{UMass}$ is based on document co-occurrence counts, a one-preceding segmentation and a logarithmic conditional probability as confirmation measure [19]. The main idea of this coherence is that the occurrence of every top word should be supported by every top preceding top word. The probabilities are derived using document co-occurrence counts. The single conditional probabilities are summarized using the arithmetic mean. $C_V$ is based on a sliding window, a one-set segmentation of the top words and an indirect confirmation measure that uses NPMI and the cosine similarity [28]. This coherence measure retrieves co-occurrence counts for the given words using a sliding window and the window size 110. The coherence is the arithmetic mean of these similarities. $C_A$ is based on a context window, a pairwise comparison of the top words and an indirect confirmation measure that uses NPMI and the cosinus similarity [28]. This coherence measure retrieves co-occurrence counts for the given words using a context window with the window size 5. $C_P$ is a based on a sliding window, a one-preceding segmentation of the top words and the confirmation measure of Fitelson's coherence [7]. Word co-occurrence counts for the given top words are derived using a sliding window and the window size 70.

Instead of using the collection itself to measure word association — which could reinforce noise or unusual word statistics [22] — we use a large external text data source: an English Wikipedia reference corpus of 8 million documents and all experiments are conducted on Palmetto platform [4]. The experimental results are given in Table 1. From the results we observe that that the traditional modeling method (LDA) cannot improve the performance of short text topic model. Additionally, we observe that PTM, BTM, LFTM and SATM are almost at the same level. The performance gap among the four is slightly behind LeadLDA (L-LDA) and not significant. Recall, LeadLAD uses labelled messages to help identify potential topical words. CSATM outperforms all baseline models in most cases. More importantly, **CSATM is competitive against LeadLDA, but doesn't require model training with labelled comments, which saves time and effort.**

TABLE I. AVERAGED COHERENCE, MEASURED BY DIFFERENT METHODS. THE TOP TWO RESULTS ARE IN BOLDFACE AND ITALIC RESPECTIVELY

| Measure | Cv | Cp | Cuci | Cumass | Cnpmi | Ca |
|---|---|---|---|---|---|---|
| LDA | 0.3899 | -0.01367 | -1.4554 | -4.1855 | -0.0373 | 0.13653 |
| PTM | 0.3671 | *0.07687* | *-0.9579* | **-2.7827** | -0.0215 | 0.0907 |
| BTM | 0.3719 | 0.01453 | -1.1233 | -3.0080 | -0.0215 | 0.15107 |
| L-LDA | **0.3957** | 0.0539 | -1.0948 | -2.9621 | *0.0175* | *0.1533* |
| LFTM | 0.3587 | 0.0443 | -2.0121 | -3.0382 | 0.0065 | 0.0887 |
| SATM | 0.3681 | 0.0322 | -1.0862 | -3.1637 | 0.0074 | 0.1112 |
| CSATM | *0.3903* | **0.0792** | **-0.9154** | *-2.8259* | **0.0201** | **0.16553** |

### C. Topic Assignment Evaluation

After extracting high-quality topics from the corpus, the assignments of topics to comments should have reasonable accuracy; sometimes it is important to know the "targets" each comment discusses in some downstream applications like stance detection, opinion mining, and so on. In our experiment, we labelled the topic assignments to the top 100 comments in each discussion thread, and compared the performance on CSATM to other models in terms of the accuracy of topic assignment, and the results are given in Fig. 6.

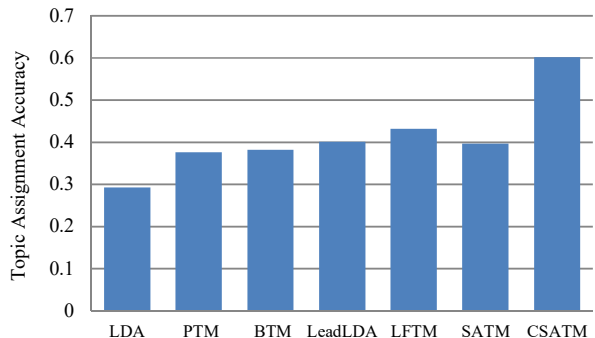

Figure 6. Accuracy of topic assignments to comments

We observe that CSATM achieves much higher accuracy than other models. That's because conventional models cannot deal with noisy comments like emojis, pictures, cursing, and so on in online discussions. CSATM has the ability to find the correct topic distributions of comments through their ancestors in the discussion thread using the proposed topic transitivity property. Take the discussion thread in Fig. 1 as an example, there are two topics in that discussion: "concept completely blows your mind" and "all roads work by being connected up". Topics to all comments may be correctly assigned except comment node 9 that is an emoji. Traditional models may fail to assign the right topic for this comment and randomly pick up one. Our model can make the topic of comment 9 correctly assigned by inferring its background information through the conversational structure.

The accuracy of CSATM is still below 0.6 because some of the topics discovered are not correct, so the assignments of topics to comments make no sense in this case. The assignment error of comments leading discussions will affect the correctness of topic assignments of their dependents.

## V. CASE STUDY

In this section, we use a real case as demo to show the effectiveness of our model. The left box in Fig. 6 is a snippet of an online discussion on the news "Texas serial bomber made video confession before blowing himself up". Topics are bolded and marked by different colors. We can see there are basically three topics discussed in this thread: 1. the news title, 2. chance to see the video, 3. Browns win the Super Bowl. This is a very typical and special case, because the topical words are very sparse, and one topic (browns win super bowl) shifts from the main discussion thread.

We set the number as three and use four different topic models to extract the topics: LDA, PTM, BTM and CSATM. We can see that LDA extracted topic 2 and 3, but they are mixed together. PTM extracted topic 2 and 3, but did not capture enough topical words for topic 3. BTM only extracted topic 2. All the three models failed to extract topic 1. Compared to the above three models, CSATM shows great performance by successfully extracted all the three topics with enough topical words. For topics that lead the discussions but their topical words are not repeatedly occurred in the comments and replies, conventional topic models based on word occurrence may not extract such kind of topics successfully, but our proposed model CSATM could deal with this issue. Of course, when the data is not sparse and topic word occurrence is high enough for modeling, CSATM can also achieve good performance by setting the difference of the weight sequence in equation (1) to a smaller to value until 1.

## VI. CONCLUSION AND FUTURE WORKS

In this paper, we have proposed the topic "popularity" and "transitivity" intuitions and presented a novel topic model CSATM for online discussions. Conventional works considering only plain text streams are not sufficient enough to summarize noisy discussion trees. CSATM captures the conversational structure as context for topic modelling and topic assignment to each comment, leading to better performance in terms of topic coherence and assignment accur-acy. By comparing our proposed model with a number of state-of – the –art baseline models on real word datasets, we have demonstrated competitive results, and the effectiveness of using conversational discourse structure to help in identifying topical content embedded in short and colloquial online discussions. Weight sequence selection may be a little confusing, but that is due to the inherent subjectivity of topic modeling and there are no uniform standards for measure a topic good or not even its coherence score is high enough. In future work, we will explore and explain this part more.

---
[4] http://aksw.org/Projects/Palmetto.html

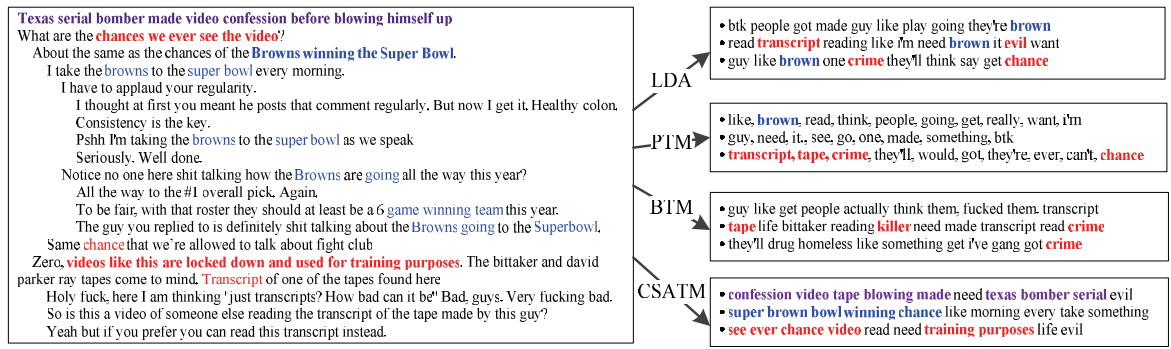

Figure 7. An example thread of user comments on the news: " Texas serial bomber made video confession before blowing himself up"[5] Three topics are bolded and marked by different colors

There are plentiful downstream applications of our proposed model. For example, it can assistant users to browse a long discussion thread quickly by summarizing the possible topics. Oftentimes, a popular news article or interesting post can easily accumulate thousands of comments within a short period of time, which makes it difficult for interested users to access and digest information in such data. Therefore, modeling the user-generated comments with respect to different topics and automatically gaining the insight of readers' opinions and attention on the news event will save users' a lot of time.

It will also very helpful for sentiment analysis or stance detection. The massive amount of online discussions provides us with valuable resources for studying and understanding public opinions on fundamental societal issues, e.g., abortion or gun rights. Automatically predicting user stance and identifying corresponding arguments are important tasks for improving policy-making process and public deliberation. Traditional stance detection methods assume that there is only one topic in a discussion and try to classify the stance into positive and negative. However, there are sometimes multiple topics within one discussion thread. For each of the topics, people have various stances, so it will be more fine-grained if we can classify users' sentiments according to topics. More downstream applications will be explored in the future.

ACKNOWLEDGMENT

This work was supported by the Ohio Department of Higher Education, the Ohio Federal Research Network and the Wright State Applied Research Corporation under award WSARC-16-00530 (C4ISR: Human-Centered Big Data).